\documentclass[10pt,twocolumn,letterpaper]{article}
\usepackage{iccv}
\usepackage{times}
\usepackage{epsfig}
\usepackage{bigstrut}
\usepackage{bm}
\usepackage{enumerate}
\usepackage{graphicx}
\usepackage{amsmath}
\usepackage{multirow}
\usepackage{amssymb}
\usepackage{color}
\usepackage{array}
\usepackage{float}
\usepackage[caption=false,font=footnotesize]{subfig}
\graphicspath{{figures/}}

\usepackage[pagebackref=true,breaklinks=true,letterpaper=true,colorlinks,bookmarks=false]{hyperref}

\newcommand{\comment}[1]{}
\newcommand{\deflen}[2]{%
    \expandafter\newlength\csname #1\endcsname
    \expandafter\setlength\csname #1\endcsname{#2}%
}

\iccvfinalcopy 

\def\iccvPaperID{} 

\ificcvfinal\fi
\begin{document}

\title{Cascade Residual Learning: A Two-stage Convolutional Neural\\Network for Stereo Matching}

\author{Jiahao Pang$^*$\hspace{15pt}Wenxiu Sun\thanks{Both authors contributed equally.}\hspace{18pt}Jimmy SJ. Ren\hspace{15pt}Chengxi Yang\hspace{15pt}Qiong Yan\\
SenseTime Group Limited\\
{\tt\small \{pangjiahao, sunwenxiu, rensijie, yangchengxi, yanqiong\}@sensetime.com}
}

\maketitle
\thispagestyle{empty}

\begin{abstract}
Leveraging on the recent developments in convolutional neural networks (CNNs), matching dense correspondence from a stereo pair has been cast as a learning problem, with performance exceeding traditional approaches. 
However, it remains challenging to generate high-quality disparities for the inherently ill-posed regions. 
To tackle this problem, we propose a novel cascade CNN architecture composing of two stages. 
The first stage advances the recently proposed DispNet by equipping it with extra up-convolution modules, leading to disparity images with more details. 
The second stage explicitly rectifies the disparity initialized by the first stage; it couples with the first-stage and generates residual signals across multiple scales. 
The summation of the outputs from the two stages gives the final disparity. 
As opposed to directly learning the disparity at the second stage, we show that residual learning provides more effective refinement. 
Moreover, it also benefits the training of the overall cascade network. 
Experimentation shows that our cascade residual learning scheme provides state-of-the-art performance for matching stereo correspondence. 
By the time of the submission of this paper, our method ranks first in the KITTI 2015 stereo benchmark, surpassing the prior works by a noteworthy margin.
\end{abstract}


\section{Introduction}
\label{sec:intro}
Dense depth data is indispensable for reconstructing or understanding a 3D scene. 
Although active 3D sensors such as Lidar, ToF, or structured light can be employed, sensing depth from stereo cameras is typically a more cost-effective approach. 
Given a rectified stereo pair, depth can be estimated by matching corresponding pixels on the two images along the same scan-line.
Particularly, for an arbitrary pixel $(x, y)$ in left image, suppose its correspondence is found at location $(x+d, y)$ in the right image, we can compute its depth by $f\bm\cdot l/d$, where $f$ is the focal length, $l$ is the baseline distance, and $d$ is often referred to as disparity. 
As depth is inversely proportional to disparity, a stereo matching system is targeted to produce an accurate dense disparity instead. 

Stereo matching is traditionally formulated as a problem with several stages of optimization. 
Until recent years with the developments in convolutional neural networks~(CNNs) \cite{lecun15}, it is cast as a learning task. 
Taking advantages of the vast available data, correspondence matching with CNNs achieves considerable gain compared to traditional approaches in terms of both accuracy and speed. 
Nevertheless, it is still difficult to find the correct correspondence at inherently ill-posed regions, such as object occlusions, repeated patterns, or textureless regions. 
For a pixel appearing in one image yet occluded in the other, its correspondence cannot be identified; while for repeated patterns and textureless regions, many potential correspondences exists. 
All these issues lead to erroneous disparity estimations.

To alleviate the aforementioned problems, we propose a 
{\it cascade residual learning} (CRL) framework, composing of two stages of convolutional neural networks with hour-glass structure\,\cite{dosovitskiy15, ilg17}. 
At the first-stage network, an simple-yet-nontrivial up-convolution module is introduced to produce \emph{fine-grained} disparities, setting up a good starting point for the residual learning at the second stage.
At the second stage, the disparity is explicitly rectified with the residual signals produced at multiple scales.
It is easier to learn the residual than to learn the disparity directly, similar to the mechanism of ResNet \cite{he16}. To the extreme where the initial disparity is already optimal, the second-stage network can simply generate zero residual to keep the optimality. 
However, the building blocks of \cite{he16}---residual blocks---are cascaded one-by-one, in which the residuals cannot be directly supervised. 
Different from \cite{he16} and other works along its line ({\it e.g.}, \cite{carreira16}), we embed the residual learning mechanism across multiple scales, where the residuals are explicitly supervised by the difference between the ground-truth disparity and the initial disparity, leading to \emph{superior} disparity refinement.

The proposed CRL scheme is trained end-to-end, integrating the traditional pipeline\,\cite{scharstein02} from matching cost computation, cost aggregation, disparity optimization, to disparity refinement by a stack of non-linear layers. 
The two stages of CRL boost the performance together and achieve state-of-the-art stereo matching results. 
It \emph{ranks first} in the KITTI 2015 stereo benchmark \cite{menze15}.


Our paper is structured as follows. 
We review related works in Section\,\ref{sec:related}. 
Then we elaborate our CRL framework and discuss our network architecture in Section\,\ref{sec:method}. 
In Section\,\ref{sec:results} and Section\,\ref{sec:conclude}, experimentation and conclusions are presented respectively.

\section{Related Works}
\label{sec:related}
There exist a large body of literature on stereo matching. 
We hereby review a few of them, with emphasis placed on those recent methods employing convolutional neural networks (CNNs). 

A typical stereo matching algorithm, {\it e.g.,}~\cite{hirschmuller08,sun03}, consists of four steps~\cite{scharstein02}: 1) matching cost computation; 2) cost aggregation; 3) disparity optimization (derive the disparity from the cost volume); and 4) disparity refinement (post-process the disparity). 
In contrast, CNN-based approaches estimate disparities reflecting part or all of the aforementioned four steps. 
These approaches can be roughly divided into the three categories.

\textbf{Matching cost learning}: In contrast to hand-crafted matching cost metrics, such as sum of absolute difference (SAD), normalized cross correlation (NCC) and Birchfield-Tomasi cost~\cite{birchfield98}, CNNs are utilized to measure the similarity between image patches. 
Han~{\it et al.}~\cite{han15} presented a Siamese network called MatchNet, which extracts features from a pair of patches followed by a decision module for measuring similarity. 
Concurrently, Zagoruyko~{\it et al.}~\cite{zagoruyko15} and Zbontar~{\it et al.}~\cite{zbontar16} investigated a series of CNN architectures for binary classification of pairwise matching and applied in disparity estimation. 
In contrast to an independent binary classification scheme between image patches, Luo~{\it et al.}~\cite{luo16} proposed to learn a probability distribution over all disparity values. 
This strategy employs a diverse set of training samples without concerning about the unbalanced training samples. 
Though the data-driven similarity measurements out-perform the traditional hand-crafted ones, a number of post-processing steps ({\it e.g.,} steps 2) to 4) in the traditional stereo matching pipeline) are still necessary for producing compelling results.

\textbf{Regularity learning}: Based on the observation that disparity images are generally piecewise smooth, some existing works imposes smoothness constraints in the learning process.
Menze~{\it et al.}~\cite{menze15} applied adaptive smoothness constraints using texture and edge information for a dense stereo estimation. 
By discovering locally inconsistent labeled pixels, Gidaris~{\it et al.}~\cite{gidaris17} propose the detect, replace, refine framework. 
However, discarding unreliable disparities with new ones results in a wasted computation.
Disparity can also be regularized by incorporating with mid- or high-level vision tasks. 
For instance, disparity was estimated concurrently by solving the problem of semantic segmentation, {\it e.g.}, \cite{bleyer11, ladicky12, yamaguchi14}. Guney and Geiger raised Displets in \cite{guney15}, which utilizes object recognition and semantic segmentation for finding stereo correspondence.

\textbf{End-to-end disparity learning}: By carefully designing and supervising the network, a fine disparity is able to be end-to-end learned with stereo inputs. Mayer~{\it et al.}~\cite{mayer16} presented a novel approach called DispNet, where an end-to-end CNN is trained using synthetic stereo pairs. In parallel with the proposal of DispNet, similar CNN architectures are also applied to optical flow estimation, leading to FlowNet~\cite{dosovitskiy15} and its successor, FlowNet\,2.0~\cite{ilg17}. A very recent method, GC-NET\,\cite{kendall17}, manages to employ contextual information with 3D convolutions for learning disparity. 
For monocular depth estimation, end-to-end semi-supervised~\cite{kuznietsov17} and unsupervised~\cite{godard17} approaches were also proposed, which connect stereo images with the estimated disparity and require a very limited amount of training data that has ground-truth disparity.

Our work belongs to the third category. In spite of the superior performance of the CNN-based approaches, it remains very challenging to produce accurate disparities at ill-posed regions. Unlike existing works, we present a cascade residual learning scheme to tackle the aforementioned problem. Particularly, we adopt a two-stage CNN, in which the first stage delivers a high-quality initialized disparity map. After that, the second stage performs further refinement/rectification by producing residual signals across multiple scales. Our experimentation shows that, the proposed cascade residual learning scheme provides state-of-the-art disparity estimates with an acceptable runtime.




\section{Cascade Residual Learning}
\label{sec:method}
\begin{figure*}[t]
\centering
    \includegraphics[height=110pt]{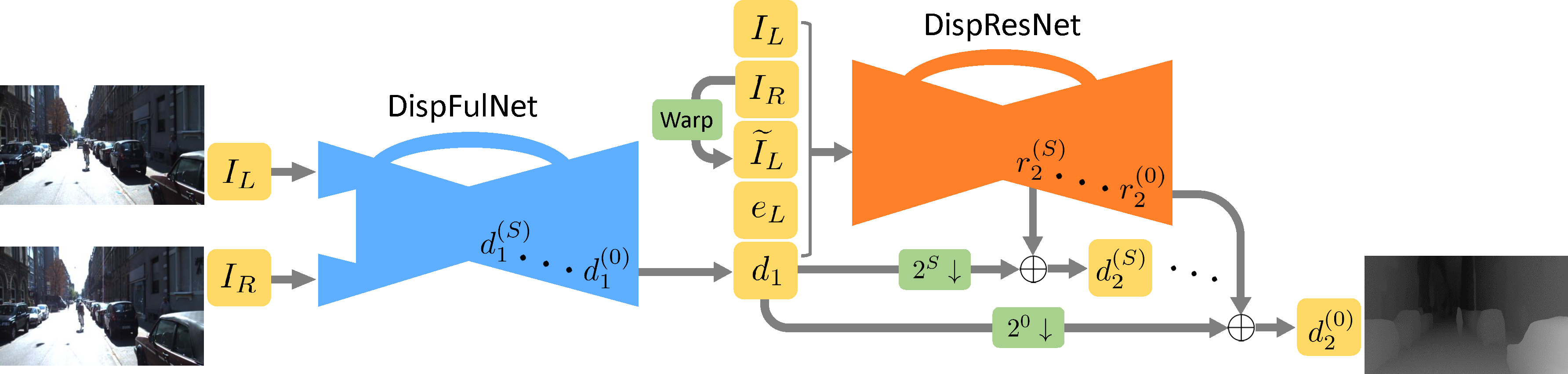}
    \vspace{5pt}
\caption{Network architecture of our \emph{cascade residual learning} (CRL) scheme. The first stage is \emph{DispFulNet} and the second stage is \emph{DispResNet} with multiscale residual learning. The module ``\,$2^{s}\hspace{-3pt}\downarrow$'' is the downsampling layer to shrink $d_1$ for $2^s$ times, while ``Warp'' denotes the warping layer.}
\label{fig:arch}
\end{figure*}

This section illustrates our cascade residual learning (CRL) scheme in detail. 
\subsection{Two-stage Disparity Computation}
In general, low-level vision tasks, {\it e.g.}, denoising and deblurring, can be improved with post-facto iterative refinement \cite{milanfar13}, and disparity/flow estimation is no exception \cite{brox11}. 
Recently, Ilg~{\it et al.}~\cite{ilg17} introduced FlowNet\,2.0, which uses stacking CNNs for optical flow refinement and achieves reasonable gain. 
The lessons of the previous works inspire us to employ a two-stage CNN for disparity estimation.

Akin to the proposal of DispNetC (``C'' indicates the network has a correlation layer) \cite{mayer16}, the first stage of our CNN has an hour-glass structure with skip connections. 
However, DispNetC outputs disparity image at half the resolution of the input stereo pair. 
Differently, our network includes extra deconvolution modules to magnify the disparity, leading to disparity estimates at the same size of the input images. 
We call our first stage network \emph{DispFulNet} (``Ful'' means full-resolution). 
As shown later in Section\,\ref{sec:results}, our DispFulNet provides extra details and sharp transitions at object boundaries, serving as an ideal starting point for the second-stage refinement.

Note that in our network, the two stages are cascaded in a way recommended by \cite{ilg17}. 
Specifically, the first network takes as input the stereo pair $I_L$ and $I_R$ and produces the initial disparity $d_1$ (of the left image). 
We then warp the right image $I_R$ according to disparity $d_1$ and obtain a synthesized left image, {\it i.e.},
\begin{equation}
    \widetilde{I}_L(x,y) = I_R(x + d_1(x,y), y).
\end{equation}
Then the input to the second network is the concatenation of $I_L$, $I_R$, $d_1$, $\widetilde{I}_L(x,y)$ and the error $e_L = |I_L-\widetilde{I}_L(x,y)|$. 
The warping operation is differentiable for bilinear interpolation \cite{ilg17,jaderberg15}, hence our network can be trained end-to-end.
\subsection{Mutiscale Residual Learning}
For the second-stage refinement/rectification, we propose to adopt the residual learning scheme of He~{\it et al.}~\cite{he16}.
Particularly, given the initial disparity $d_1$ obtained with the first stage, the second network outputs the corresponding residual signal $r_2$, then the new disparity $d_2$ is given by $d_1 + r_2$. 
In this way, we relieve the ``burden'' of the second-stage network, letting it only focus on learning the highly nonlinear residual. 
On par with the spirit in \cite{he16}, in the extreme case when the first stage already produces the optimal disparity, the second-stage network only needs to output zero residual to retain the optimality.

The second-stage of our architecture also takes an hour-glass structure, producing residual signals across multiple scales.
We call our second-stage network \emph{DispResNet} (``Res'' means residual). 
In the expanding part of DispResNet, the residuals are produced across several scales. 
They are denoted as $\{r_2^{(s)}\}_{s=0}^S$ where $0$ denotes the scale of full resolution. 
The summation of $r_2^{(s)}$ with the downsampled disparity $d_1^{(s)}$ leads to the new disparity at scale $s$, {\it i.e.},
\begin{equation}
    d_2^{(s)} = d_1^{(s)} + r_2^{(s)}, 0\le s\le S.
\end{equation}
To train DispResNet, we supervise the estimated disparities $\{d_2^{(s)}\}_{s=0}^S$ across $S+1$ scales. Hence, differs from the off-the-shelf residual block structure proposed in \cite{he16}, our network explicitly supervise the residual signals, leading to effective disparity refinement.

In fact, a straightforward application of FlowNet\,2.0\,\cite{ilg17} for disparity estimation is to adopt DispNetS\,\cite{mayer16}---a variation of DispNetC without correlation layer and ``S'' means simple---to \emph{directly} learn the disparity. 
Nevertheless, our comparisons in Section\,\ref{sec:results} show that incorporating residual learning brings more gain than its direct learning counterpart, {\it i.e.}, DispNetS. 
Furthermore, residual learning also benefits the finetuning of the overall network, as it alleviates the problem of over-fitting~\cite{he16,ilg17}, while using DispNetS harms the performance after overall finetuning.

\subsection{Network Architecture}
Our CRL architecture is illustrated in Fig.\,\ref{fig:arch}, where $d_1=d_1^{(0)}$, and the final disparity output is $d_2^{(0)}$. To obtain the downsampled disparity images $\{d_1^{(s)}\}_{s=0}^{S}$, we have implemented a differentiable bilinear downsampling layer, similar to the sampler module in the spatial transformer networks\,\cite{jaderberg15}.

The first stage, DispFulNet, enlarges the half-resolution disparity estimates of DispNetC \cite{mayer16}.
For a concise presentation, the detailed architecture of DispFulNet is not provided here. 
In general, it shares similar spirits with DispNetC. 
Though differently, we append extra up-convolutions to the last two convolution layers of DispNetC, the output of the upconvolutions are then concatenated with the left image. 
By applying one more convolution (with one output channel) to the concatenated 3-D array, we arrive at the output of DispFulNet---a full-resolution disparity image. 
The full-resolution disparity image, along with the other intermediate disparity images at six different scales, are supervised by the ground-truth through computing the $\ell_1$ loss.

The detailed specification of the second stage, DispResNet, is provided in Table.\,\ref{tab:dispres}. 
Note that at a certain scale, say, $1/4$, the bilinear downsampling layer {\tt pr\_s1\_4} shrinks {\tt pr\_s1}, the disparity prediction of DispFulNet, by a factor of 4. 
The downsampled disparity is then added to the learned residual {\tt res\_4} by the element-wise summation layer {\tt pr\_s2\_4}, leading to the disparity prediction at scale $1/4$. 
We follow the typical supervised learning paradigm and compute an $\ell_1$ loss between the disparity estimate and the ground-truth disparity at each scale.

\begin{table}[t]
  \centering\scriptsize
  \begin{tabular}{|m{22pt}<{\centering}|m{5pt}<{\centering}m{5pt}<{\centering}m{30pt}<{\centering}|m{5pt}<{\centering}m{5pt}<{\centering}|m{75pt}<{\centering}|}
    \hline
    \textbf{Layer} & \textbf{K} & \textbf{S} & \textbf{Channels} & \textbf{I} & \textbf{O} & \textbf{Input Channels}\bigstrut\\
    \hline
    conv1 & 5     & 1     & 13/64 & 1     & 1     & left+right+left\_s+err+pr\_s1 \bigstrut[t]\\
    conv2 & 5     & 2     & 64/128 & 1     & 2     & conv1 \\
    conv2\_1 & 3     & 1     & 128/128 & 2     & 2     & conv2 \\
    conv3 & 3     & 2     & 128/256 & 2     & 4     & conv\_3\_1 \\
    conv3\_1 & 3     & 1     & 256/256 & 4     & 4     & conv3 \\
    conv4 & 3     & 2     & 256/512 & 4     & 8     & conv3\_1 \\
    conv4\_1 & 3     & 1     & 512/512 & 8     & 8     & conv4 \\
    conv5 & 3     & 2     & 512/1024 & 8     & 16    & conv4\_1 \\
    conv5\_1 & 3     & 1     & 1024/1024 & 16    & 16    & conv5 \bigstrut[b]\\
    \hline
    \hline
    res\_16 & 3     & 1     & 1024/1 & 16    & 16    & conv5\_1 \bigstrut[t]\\
    pr\_s1\_16 & -   & -   & 1/1   & 1     & 16    & pr\_s1 \\
    pr\_s2\_16 & -   & -   & 1/1   & 16    & 16    & pr\_s1\_16+res\_16 \bigstrut[b]\\
    \hline
    upconv4 & 4     & 2     & 1024/512 & 16    & 8     & conv5\_1 \bigstrut[t]\\
    iconv4 & 3     & 1     & 1025/512 & 8     & 8     & upconv4+conv4\_1+pr\_s2\_16 \\
    res\_8 & 3     & 1     & 512/1 & 8     & 8     & iconv4 \\
    pr\_s1\_8 & -   & -   & 1/1   & 1     & 8     & pr\_s1 \\
    pr\_s2\_8 & -   & -   & 1/1   & 8     & 8     & pr\_s1\_8+res\_8 \bigstrut[b]\\
    \hline
    upconv3 & 4     & 2     & 512/256 & 8     & 4     & iconv4 \bigstrut[t]\\
    iconv3 & 3     & 1     & 513/256 & 4     & 4     & upconv3+conv3\_1+pr\_s2\_8 \\
    res\_4 & 3     & 1     & 256/1 & 4     & 4     & iconv3 \\
    pr\_s1\_4 & -   & -   & 1/1   & 1     & 4     & pr\_s1 \\
    pr\_s2\_4 & -   & -   & 1/1   & 4     & 4     & pr\_s1\_4+res\_4 \bigstrut[b]\\
    \hline
    upconv2 & 4     & 2     & 256/128 & 4     & 2     & iconv3 \bigstrut[t]\\
    iconv2 & 3     & 1     & 257/128 & 2     & 2     & upconv2+conv2\_1+pr\_s2\_4 \\
    res\_2 & 3     & 1     & 128/1 & 2     & 2     & iconv2 \\
    pr\_s1\_2 & -   & -   & 1/1   & 1     & 2     & pr\_s1 \\
    pr\_s2\_2 & -   & -   & 1/1   & 2     & 2     & pr\_s1\_2+res\_2 \bigstrut[b]\\
    \hline
    upconv1 & 4     & 2     & 128/64 & 2     & 1     & iconv2 \bigstrut[t]\\
    res\_1 & 5     & 1     & 129/1 & 1     & 1     & upconv1+conv1+pr\_s2\_2 \\
    pr\_s2 & -   & -   & 1/1   & 1     & 1     & pr\_s1+res\_1 \bigstrut[b]\\
    \hline
  \end{tabular}%
  \vspace{10pt}
  \caption{Detailed architecture of the proposed \emph{DispResNet}. Layers with prefix {\tt pr\_s1} are downsampling layers applying on the predictions of the first stage; while layers with prefix {\tt pr\_s2} are element-wise summation layers leading to predictions of the second stage. {\bf K} means kernel size, {\bf S} means stride, and {\bf Channels} is the number of input and output channels. {\bf I} and {\bf O} are the input and output downsampling factor relative to the input. The symbol {\tt +} means summation for element-wise summation layers; otherwise it means concatenation.}
  \label{tab:dispres}%
\end{table}%

One may raise a straightforward question about our design: if a two-stage cascade architecture performs well, why not stacking more stages? 
First, adding more stages translates to higher computational cost and memory consumption, which is unrealistic for many practical applications. 
Second, in this paper, we aim at developing a two-stage network, where the first one manages to produce full-resolution initializations; while the second stage tries its best to refine/remedy the initial disparities with residual learning. 
The two stages play their own roles and couple with each other to provide satisfactory results. 
As to be seen in Section\,\ref{ssec:res_oth}, our two-stage network estimate high-quality disparity images with an acceptable execution time: it takes 0.47\,sec with an Nvidia GTX 1080 GPU to obtain a disparity image in the KITTI 2015 stereo dataset.


\section{Experiments}
\label{sec:results}
Experimental setup and results are presented in this section. 
To evaluate the effectiveness of our design, we replace the two stages of our network with the plain DispNetC and/or DispNetS \cite{mayer16} for comparisons. 
We also compare our proposal with other state-of-the-art approaches, {\it e.g.},\,\cite{yamaguchi14,zbontar16}.
\begin{table}[t]
  \centering\small
    \begin{tabular}{|c|c|c|}
    \hline
    \multirow{2}[4]{*}{Target Dataset} & \multicolumn{2}{c|}{Training Schedule} \bigstrut\\
\cline{2-3}          & Separate & Overall \bigstrut\\
    \hline
    \hline
    FlyingThings3D & {\tt 1F-2F} & {\tt 1F-2F-0F}\bigstrut[t]\\
    Middlebury & {\tt 1F-2F} & {\tt 1F-2F-0F}  \\
    KITTI & {\tt 1F-1K-2F-2K} & {\tt 1F-1K-2F-2K-0K} \bigstrut[b]\\
    \hline
    \end{tabular}%
  \vspace{10pt}
  \caption{Training schedules of a two-stage network with different target datasets. When overall finetuning is adopted, the whole network is finetuned on the target dataset at the end.}
  \label{tab:train}%
\end{table}%

\begin{table*}[htbp]
  \centering\small
    \begin{tabular}{m{42pt}<{\centering}|m{42pt}<{\centering}||m{16pt}<{\centering}m{20pt}<{\centering}|m{16pt}<{\centering}m{20pt}<{\centering}||m{16pt}<{\centering}m{24pt}<{\centering}|m{16pt}<{\centering}m{24pt}<{\centering}||m{16pt}<{\centering}m{24pt}<{\centering}|m{16pt}<{\centering}m{20pt}<{\centering}}
    \hline
    \multicolumn{2}{c||}{Architecture} & \multicolumn{12}{c}{Dataset} \bigstrut[t]\\
    \hline
    \multirow{2}[3]{*}{Stage 1} & \multirow{2}[3]{*}{Stage 2} & \multicolumn{4}{c||}{FlyingThings3D} & \multicolumn{4}{c||}{Middlebury 2014} & \multicolumn{4}{c}{KITTI 2015} \bigstrut[b]\\
\cline{3-14}          &       & \multicolumn{2}{c|}{Separate} & \multicolumn{2}{c||}{Overall} & \multicolumn{2}{c|}{Separate} & \multicolumn{2}{c||}{Overall} & \multicolumn{2}{c|}{Separate} & \multicolumn{2}{c}{Overall} \bigstrut\\
    \hline
    DispNetC & -     & 1.84  & 9.67 & -     & -     & 1.95  & 15.42 & -     & -     & 0.77     & \comment{4.34}3.16 & -     & - \bigstrut[t]\\
    DispNetC & DispNetS & 1.74  & 7.98 & 1.77  & 9.04 & 1.94  & 14.72 & 1.96  & 14.91 & 0.75     & \comment{3.82}2.71 & 0.78  & 2.82 \\
    DispNetC & DispResNet & 1.63  & 7.76 & 1.60  & 7.67 & 1.86  & 14.69 & 1.88  & 14.71 & 0.72     & \comment{3.69}2.66 & 0.71    & 2.63 \\
    \hline
    DispFulNet & -     & 1.75  & 8.61 & -     & -     & 1.73  & 12.82 & -     & -     & 0.73     & \comment{3.14}2.41 & -     & - \bigstrut[t]\\
    DispFulNet & DispNetS & 1.51  & 6.93 & 1.53  & 7.09 & 1.52  & 9.69 & 1.51  & 10.04 & 0.72     & \comment{2.96}2.29 & 0.73     & 2.33 \\
    DispFulNet & DispResNet & 1.35  & 6.34 & \textbf{1.32}  & \textbf{6.20} & 1.46 & 9.35 & \textbf{1.40}  & \textbf{9.13} & 0.69     & \comment{2.87}2.12 & \textbf{0.68}     & \comment{\textbf{2.86}}\textbf{2.10} \\
    \hline
    \end{tabular}%
  \vspace{10pt}
  \caption{Comparisons of our CRL architecture (DispFulNet+DispResNet) with other similar networks. In each cell, the corresponding endpoint-error (EPE) and three-pixel-error (3PE) are presented, respectively.}
  \label{tab:res_arch}%
\end{table*}%

\subsection{Experimental Settings}\label{ssec:res_setup}
{\bf Datasets: }Three publicly available datasets are adopted for training and testing in this work:
\begin{enumerate}[(i)]
    \item \emph{FlyingThings3D} \cite{mayer16}: a large scale synthetic dataset containing more than 22k synthetic stereo pairs for training and 4k for testing. We found this dataset has a few images with unreasonably large disparities ({\it e.g.}, greater than $10^3$), therefore we perform a simple screening on this dataset before using it. Particularly, for a disparity image, if more than $25\%$ of its disparity values are greater than $300$, this disparity image (and the corresponding stereo pair) is removed.
    \item \emph{Middlebury 2014} \cite{scharstein14}: a small dataset capturing various high-resolution in-door scenes, which has 23 stereo pairs with given ground-truth. We only use this dataset for testing.
    \item \emph{KITTI 2015} \cite{menze15}: a real-world dataset with dynamic street views from the perspective of a driving car. It provides 200 stereo pairs with sparse ground-truth disparities and 200 pairs for evaluation through its online leaderboard. Similar to the practice in \cite{gidaris17}, we divide its training set into a training split and a validation split, where the training split occupies $85\%$ of the data and the validation split occupies the rest.
\end{enumerate}

{\bf Training: }The Caffe framework~\cite{jia14} is used to implement our CRL scheme. Generally speaking, we first train the DispFulNet, then by fixing its weights, the DispResNet is trained. 
After that, we optionally finetune the overall network. 
Depending on the targeting dataset for testing, different training schedules are employed. 
For presentation, we hereby encode every training schedule with a string. 
A segment of such string contains two characters {\tt ND}, meaning that stage {\tt N} is trained on dataset {\tt D}, with stage {\tt 0} denotes the whole network. 
For instance, {\tt 1F-1K} means the first stage is trained on the FlyingThings3D, then it is finetuned on KITTI. 
The training schedules for the three datasets are presented in Table.\,\ref{tab:train}. 
Note that the networks trained for FlyingThings3D are directly applied on the Middlebury data (at the quarter scale). 

We adopt a batch size of 4 when training the first or the second stage, and a batch size of 2 when finetuning the overall network due to limited GPU memory. 
We employ the parameters provided in \cite{mayer16} when training the first stage or the second stage on the FlyingThings3D dataset. 
During finetuning, we train the model for 200\,K iterations; however, when the target dataset is KITTI 2015, we only optimize for 100\,K iterations to lessen the problem of over-fitting. 
Since some of the ground-truth disparities are not available for the KITTI dataset, we neglect them when computing the $\ell_1$ loss.

{\bf Testing:} We test our networks on the aforementioned datasets, with two widely used metrics for evaluation:
\begin{enumerate}[(i)]
    \item \emph{Endpoint-error} (EPE): the average Euclidean distance between the estimated disparity and the ground-truth.
    \item \emph{Three-pixel-error} (3PE): computes the percentage of pixels with endpoint error more than 3. We call it three-pixel-error in this work.
\end{enumerate}

\subsection{Architecture Comparisons}
We first compare our design with several similar network architectures. 
Particularly, we use either DispNetC or DispFulNet as the first-stage network; while at the second stage, we use either DispNetS (with direct learning) or DispResNet (with residual learning) for improving the disparity estimates. 
The plain DispNetC and DispFulNet (with only one stage) are also considered in our evaluation. For DispNetC, we adopted the model released by Dosovitskiy~{\it et al.}~\cite{dosovitskiy15}; while DispFulNet are trained in a similar manner as that in \cite{dosovitskiy15} ({\it e.g.}, with multi-scale loss functions).
During the training process, we follow the schedules shown in Table\,\ref{tab:train}, hence 20 different network models are obtained for comparisons.

Objective performance of the networks on the three datasets are presented in Table.\,\ref{tab:res_arch}. 
We have the following observations:
\begin{enumerate}[(i)]
    \item Using our DispFulNet as the first stage provides \emph{higher accuracy} compared to DispNetC.
    \item Though appending a second-stage network improves the results, our DispResNet bring \emph{extra gain} compared to DispNetS (the propposal in \cite{ilg17}).
    \item When DispNetS is served as the second stage, the performance deteriorates after overall finetuning, in accordance with \cite{ilg17}. In contrast, when DispResNet is used, overall optimization further \emph{improves} the performance in most cases (except for Middlebury which is not used for training). From \cite{he16}, that is because learning the residual is less easy to over-fit the training data, making the network more stable for overall optimization.
\end{enumerate}
As a whole, our CRL scheme (DispFulNet+DispResNet) with overall finetuning achieves the best objective qualities in all the three datasets. 
In the following, we use this network model for further comparisons.

\begin{figure*}[!t]
        \centering
        \subfloat{\includegraphics[width=\figresslf]{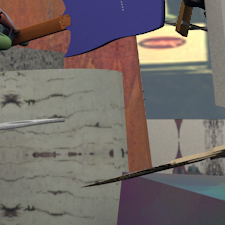}}\hspace{2pt}
        \subfloat{\includegraphics[width=\figresslf]{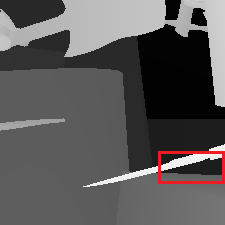}}\hspace{2pt}
        \subfloat{\includegraphics[width=\figresslf]{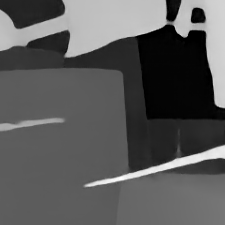}}\hspace{2pt}
        \subfloat{\includegraphics[width=\figresslf]{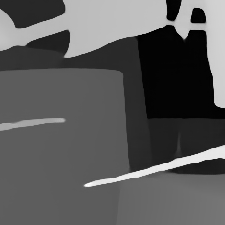}}\hspace{2pt}
        \subfloat{\includegraphics[width=\figresslf]{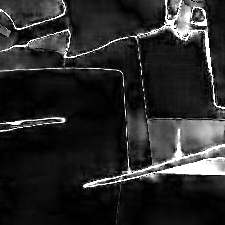}}\hspace{2pt}
        \subfloat{\includegraphics[width=\figresslf]{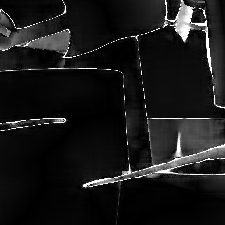}}\\ \vspace{-5pt}
        \subfloat{\includegraphics[width=\figresslf]{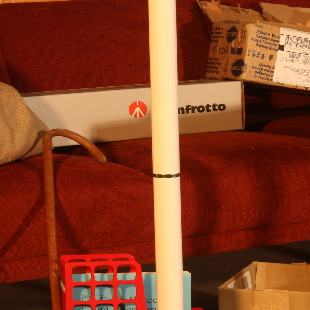}}\hspace{2pt}
        \subfloat{\includegraphics[width=\figresslf]{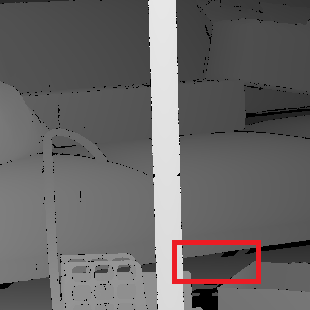}}\hspace{2pt}
        \subfloat{\includegraphics[width=\figresslf]{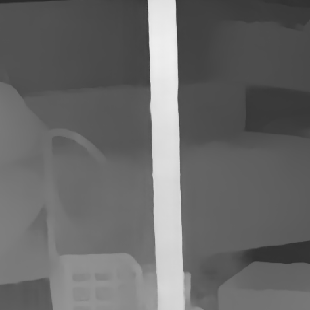}}\hspace{2pt}
        \subfloat{\includegraphics[width=\figresslf]{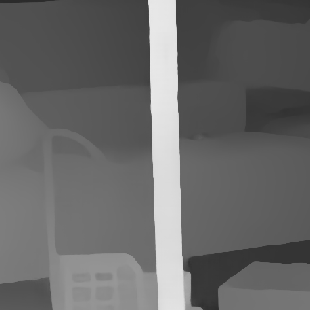}}\hspace{2pt}
        \subfloat{\includegraphics[width=\figresslf]{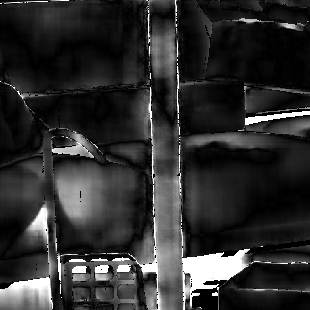}}\hspace{2pt}
        \subfloat{\includegraphics[width=\figresslf]{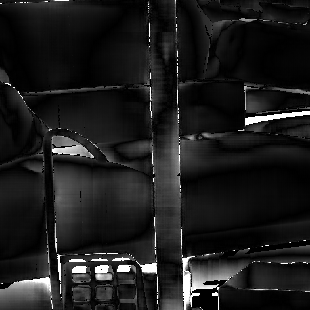}}\\
        \addtocounter{subfigure}{-10}\vspace{-5pt}
        \subfloat[Left image]{\includegraphics[width=\figresslf]{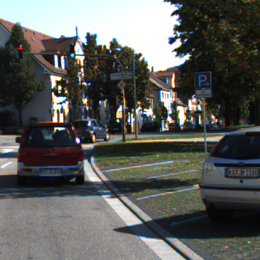}}\hspace{2pt}
        \subfloat[Ground-truth disparity]{\includegraphics[width=\figresslf]{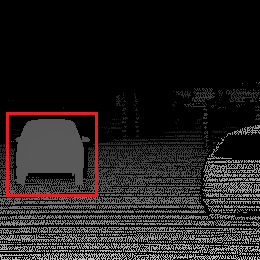}}\hspace{2pt}
        \subfloat[First-stage output]{\includegraphics[width=\figresslf]{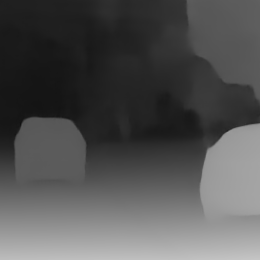}}\hspace{2pt}
        \subfloat[Second-stage output]{\includegraphics[width=\figresslf]{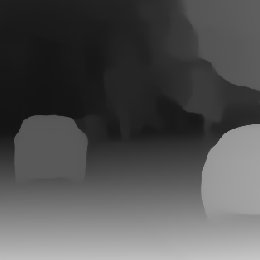}}\hspace{2pt}
        \subfloat[First-stage error]{\includegraphics[width=\figresslf]{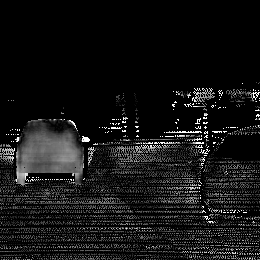}}\hspace{2pt}
        \subfloat[Second-stage error]{\includegraphics[width=\figresslf]{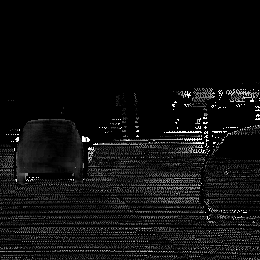}}\\
    \caption{Visual comparisons between the first-stage output by DispFulNet and the second-stage output by the whole CRL scheme (DispFulNet+DispResNet). Note that the regions within the red boxes are corrected by DispResNet.}
    \label{fig:res_slf}
\end{figure*}

Fig.\,\ref{fig:res_slf} shows the outputs of our CRL scheme and its first stage, DispFulNet, as well as their absolute differences between the ground-truth disparities. 
The three rows are segments taken from the FlyingThings3D, Middlebury and KITTI datasets, respectively. 
We see that not only the disparities at object boundaries are greatly improved by the second-stage (DispResNet), some of the occlusion and textureless regions are also rectified. 
For instance, the regions within the red boxes (on the ground-truth) are corrected by DispResNet.

\deflen{figresslf}{78pt}

\begin{figure*}[!t]
        \centering
        \subfloat{\includegraphics[width=\figresarc]{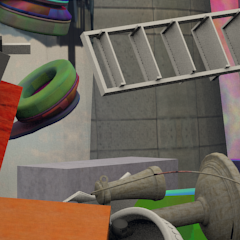}}\hspace{2pt}
        \subfloat{\includegraphics[width=\figresarc]{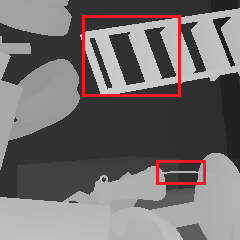}}\hspace{2pt}
        \subfloat{\includegraphics[width=\figresarc]{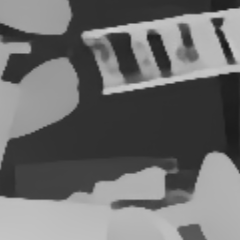}}\hspace{2pt}
        \subfloat{\includegraphics[width=\figresarc]{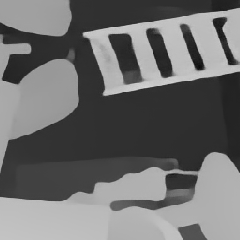}}\hspace{2pt}
        \subfloat{\includegraphics[width=\figresarc]{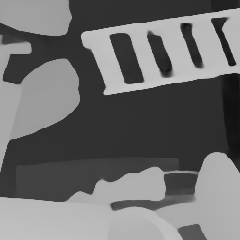}}\\ \vspace{-5pt}
        \subfloat{\includegraphics[width=\figresarc]{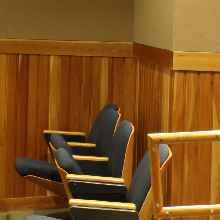}}\hspace{2pt}
        \subfloat{\includegraphics[width=\figresarc]{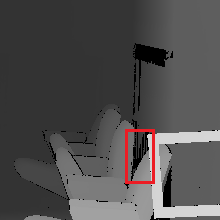}}\hspace{2pt}
        \subfloat{\includegraphics[width=\figresarc]{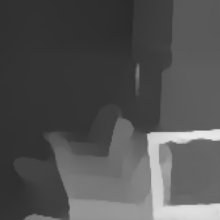}}\hspace{2pt}
        \subfloat{\includegraphics[width=\figresarc]{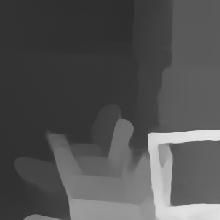}}\hspace{2pt}
        \subfloat{\includegraphics[width=\figresarc]{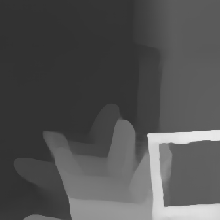}}\\
        \addtocounter{subfigure}{-10}\vspace{-5pt}
        \subfloat[Left image]{\includegraphics[width=\figresarc]{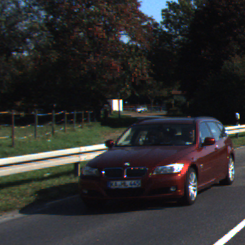}}\hspace{2pt}
            \subfloat[Ground-truth disparity]{\includegraphics[width=\figresarc]{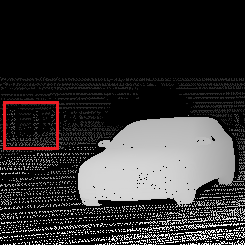}}\hspace{2pt}
        \subfloat[DispNetC+DispNetS]{\includegraphics[width=\figresarc]{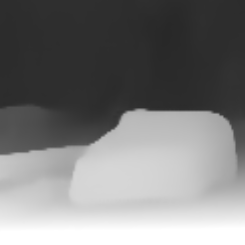}}\hspace{2pt}
        \subfloat[DispNetC+DispResNet]{\includegraphics[width=\figresarc]{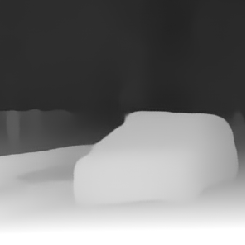}}\hspace{2pt}
        \subfloat[CRL]{\includegraphics[width=\figresarc]{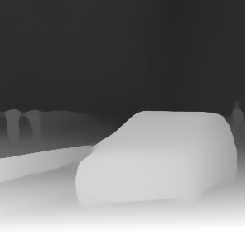}}\\
    \caption{Comparisons of three two-stage network architectures. Our proposed CRL deliveries sharpest and finest disparity images. Also note the regions bounded by the red boxes in different disparity images.}
    \label{fig:res_arch}
\end{figure*}

Fig.\,\ref{fig:res_arch} shows the disparity estimates of three different two-stage networks: DispNetC+DispNetS (akin to the proposal of \cite{ilg17}), DispNetC+DispResNet, and DispFulNet+DispResNet (our CRL), where DispNetC+DispNetS uses the model with separate training while DipsNetC+DispResNet uses the model after overall finetuning. 
Again, the three rows are segments taken from the FlyingThings3D, Middlebury and KITTI datasets, respectively. 
We see that, firstly, the proposed CRL provides sharpest disparity estimates among the three architectures, with the help of its first stage, DispFulNet. 
Furthermore, incorporating residual learning in the second stage produces high-quality disparities for ill-posed regions. 
Note the disparity estimates within the red boxes are progressively improved from DispNetC+DispNetS and DispNetC+DispResNet, to CRL.

\deflen{figresarc}{95pt}

\subsection{Comparisons with Other Methods}\label{ssec:res_oth}
In this experiment, we compare the proposed CRL to several state-of-the-art stereo matching algorithms. 
For a fair comparison, the Middlebury dataset is not adopted in this experiment as its amount of data is insufficient for finetuning our end-to-end network.

{\bf FlyingThings3D:} Since our method only takes 0.47 second to process a stereo pair in the KITTI 2015 dataset, for a fair comparison, we hereby consider three efficient yet effective methods (with code publicly available), including SPS-St\,\cite{yamaguchi14}, MC-CNN-fst\,\cite{zbontar16}, and DispNetC\,\cite{mayer16}.
We also employ the classic semi-global matching (SGM) algorithm \cite{hirschmuller08} as the baseline. 
Note that to compare with MC-CNN-fst, we train its network for 14 epochs, with a dataset containing 17 million samples extracted from the FlyingThings3D.

\begin{figure*}[!t]
        \centering
        \subfloat{\includegraphics[width=\figresmth]{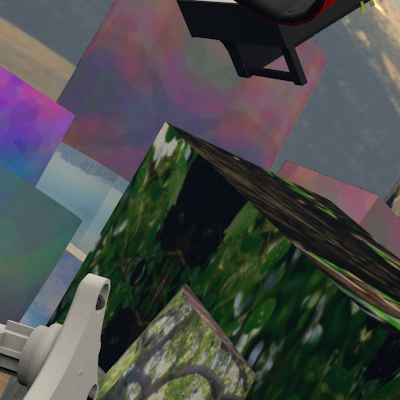}}\hspace{2pt}
        \subfloat{\includegraphics[width=\figresmth]{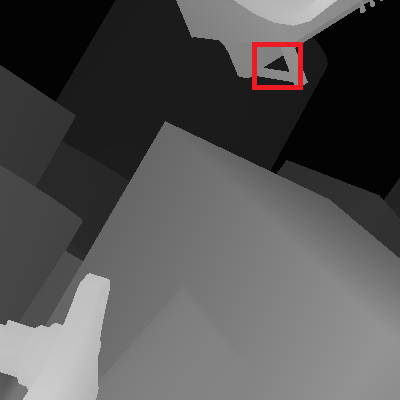}}\hspace{2pt}
        \subfloat{\includegraphics[width=\figresmth]{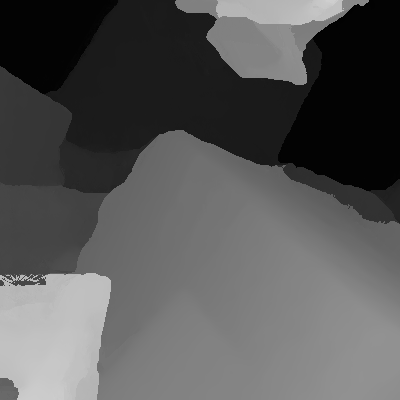}}\hspace{2pt}
        \subfloat{\includegraphics[width=\figresmth]{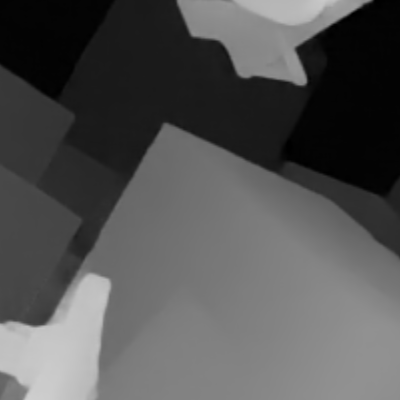}}\hspace{2pt}
        \subfloat{\includegraphics[width=\figresmth]{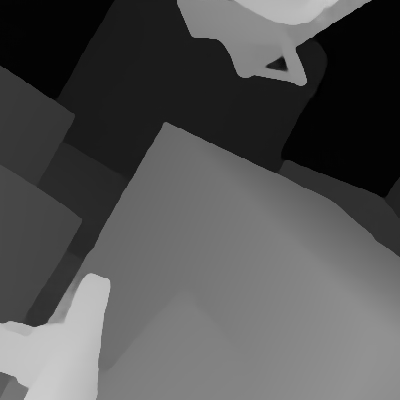}}\\ \vspace{-5pt}
        \subfloat{\includegraphics[width=\figresmth]{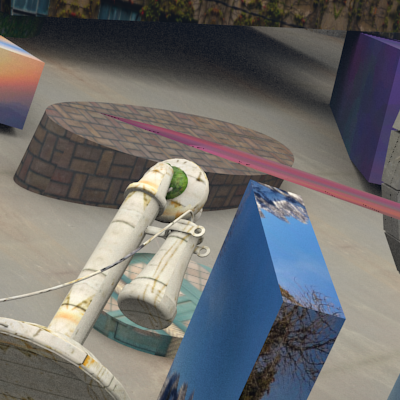}}\hspace{2pt}
        \subfloat{\includegraphics[width=\figresmth]{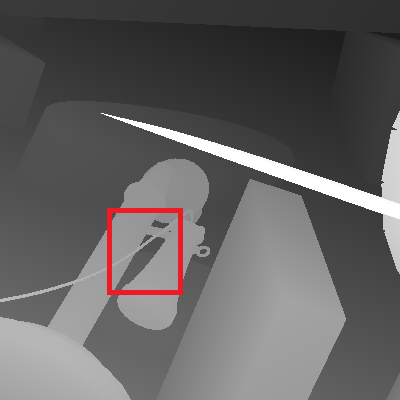}}\hspace{2pt}
        \subfloat{\includegraphics[width=\figresmth]{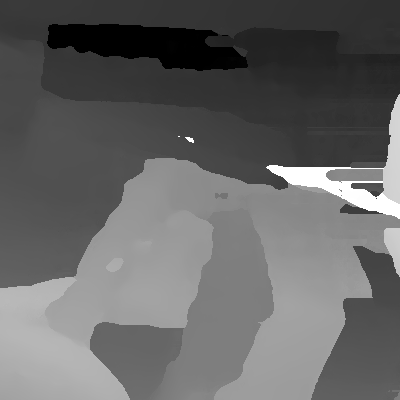}}\hspace{2pt}
        \subfloat{\includegraphics[width=\figresmth]{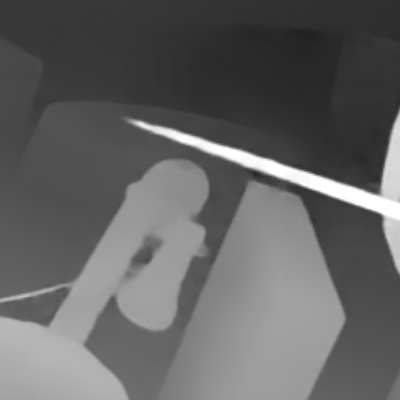}}\hspace{2pt}
        \subfloat{\includegraphics[width=\figresmth]{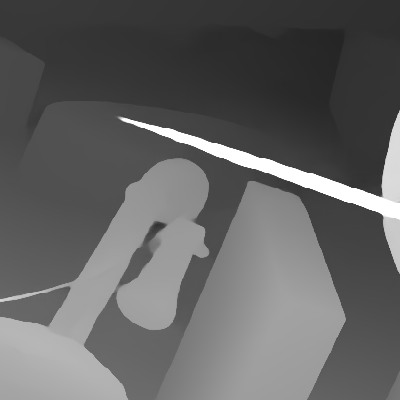}}\\ \vspace{-5pt}
        \subfloat{\includegraphics[width=\figresmth]{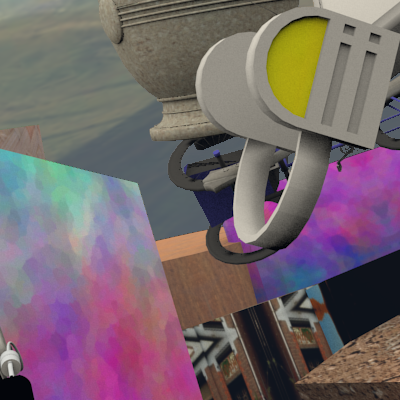}}\hspace{2pt}
        \subfloat{\includegraphics[width=\figresmth]{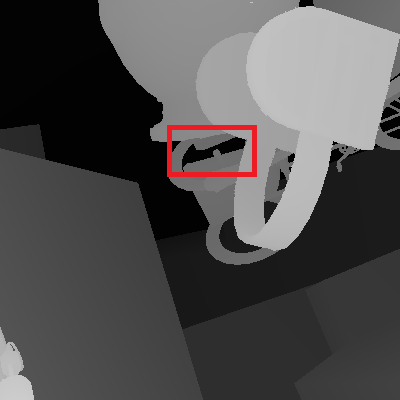}}\hspace{2pt}
        \subfloat{\includegraphics[width=\figresmth]{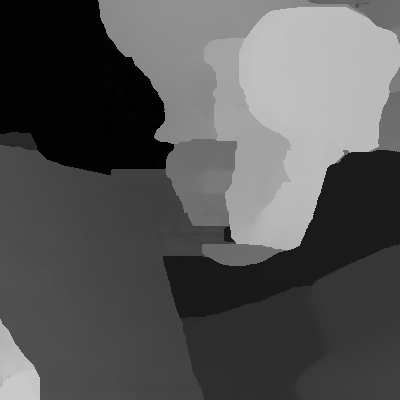}}\hspace{2pt}
        \subfloat{\includegraphics[width=\figresmth]{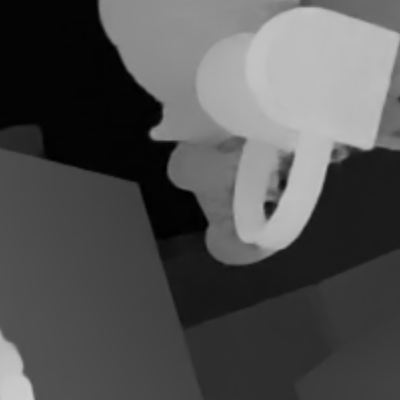}}\hspace{2pt}
        \subfloat{\includegraphics[width=\figresmth]{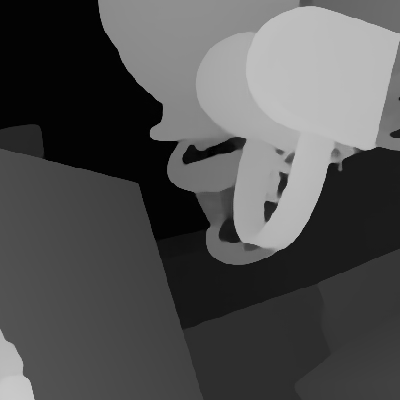}}\\
        \addtocounter{subfigure}{-10}\vspace{-5pt}
        \subfloat[Left image]{\includegraphics[width=\figresmth]{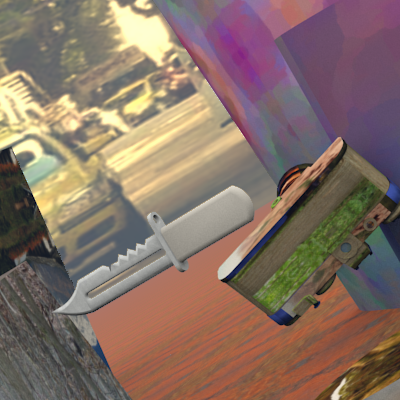}}\hspace{2pt}
        \subfloat[Ground truth disparity]{\includegraphics[width=\figresmth]{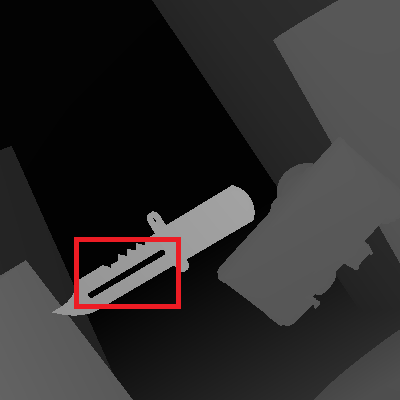}}\hspace{2pt}
        \subfloat[MC-CNN-fst]{\includegraphics[width=\figresmth]{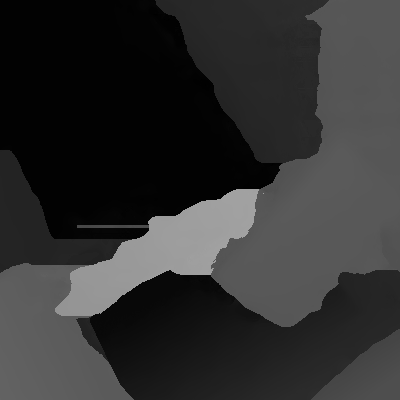}}\hspace{2pt}
        \subfloat[DispNetC]{\includegraphics[width=\figresmth]{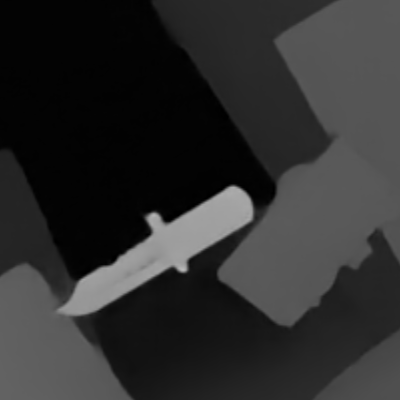}}\hspace{2pt}
        \subfloat[CRL]{\includegraphics[width=\figresmth]{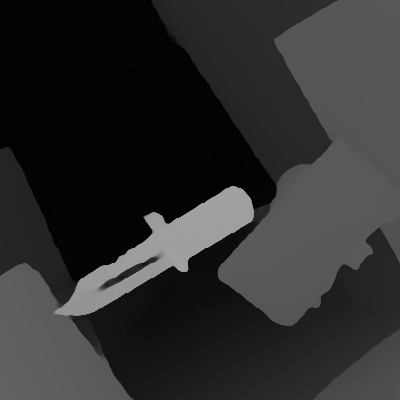}}\\
    \caption{Visual results of the proposed CRL, accompanied with those of the competing methods, on the FlyingThings3D dataset. Our method is the only one that successfully estimates the details within the red boxes.}
    \label{fig:res_mth}
\end{figure*}

Performance of the proposed CRL, along with those of the competing methods, are presented in Table.\,\ref{tab:res_fly}. 
Again, we see that our approach provides the best performance in terms of both evaluation metrics. 
In Fig.\,\ref{fig:res_mth}, we show some visual results of different approaches on the FlyingThings3D dataset, note that our CRL provides very sharp disparity estimates. Besides, our method is the only one that can generate the fine details within the red boxes.
\begin{table}[t!]
  \centering\small
    \begin{tabular}{m{22pt}<{\centering}||m{22pt}<{\centering}|m{23pt}<{\centering}|m{32pt}<{\centering}|m{35pt}<{\centering}|m{22pt}<{\centering}}
    \hline
    Metric & SGM   & SPS-St & MC-CNN-fst & \hspace{-2pt}DispNetC & CRL\\
    \hline
    EPE & 4.50  & 3.98  & 3.79  & 1.84 & 1.32\\
    3PE & 12.54 & 12.84 & 13.70 & 9.67 & 6.20\\
    \hline
    \end{tabular}%
  \vspace{10pt}
  \caption{Objective performance of our work (CRL), along with those of the competing methods on the FlyingThings3D dataset.}
  \label{tab:res_fly}%
\end{table}%

\deflen{figresmth}{93pt}

{\bf KITTI 2015 dataset:} Instead of using the training split mentioned in Section\,\ref{ssec:res_setup}, we have also trained our network on all available training data of KITTI 2015 and submitted our results to its online leaderboard. 
Table\,\ref{tab:res_kit} shows the leading submission results reported by the KITTI website, where only the three-pixel-error (3PE) values are available. 
In the table, ``All'' means all pixels are taken into account when computing 3PE, while ``Noc'' means only the non-occluded pixels are taken into account. 
The three columns ``D1-bg,'' ``D1-fg'' and ``D1-all'' means the 3PE of the background, the foreground and the all the estimates. As can be seen, our method \emph{ranks first} in the online leaderboard. 
Particularly, our overall 3PE is 2.67\%, while the second method, GC-NET\,\cite{kendall17}, has a 3PE of 2.87\%; however, our runtime is only about half of that of GC-NET. 
Visual results are not included here for conciseness, we recommend the readers go to the KITTI website\,\cite{menze15} for more details.

\begin{table*}[htbp]
\vspace{-4pt}
  \centering\small
    \begin{tabular}{rrrrrrrrr}
          &       &       &       &       &       &       &       &  \bigstrut[b]\\
\cline{2-9}          & \multicolumn{1}{c||}{\multirow{2}[4]{*}{Methods}} & \multicolumn{3}{c||}{All} & \multicolumn{3}{c||}{Noc} & \multicolumn{1}{c}{\multirow{2}[4]{*}{Runtime (sec)}} \\
\cline{3-8}          & \multicolumn{1}{c||}{} & \multicolumn{1}{c|}{\phantom{bl}D1-bg\phantom{bl}} & \multicolumn{1}{c|}{\phantom{bl}D1-fg\phantom{bl}} & \multicolumn{1}{c||}{\phantom{bl}D1-all\phantom{bl}} & \multicolumn{1}{c|}{\phantom{bl}D1-bg\phantom{bl}} & \multicolumn{1}{c|}{\phantom{bl}D1-fg\phantom{bl}} & \multicolumn{1}{c||}{\phantom{bl}D1-all\phantom{bl}} &  \\
\cline{2-9}          & \multicolumn{1}{c||}{CRL (Ours)} & \multicolumn{1}{c|}{2.48} & \multicolumn{1}{c|}{\textbf{3.59}} & \multicolumn{1}{c||}{\textbf{2.67}} & \multicolumn{1}{c|}{2.32} & \multicolumn{1}{c|}{\textbf{3.12}} & \multicolumn{1}{c||}{\textbf{2.45}} & \multicolumn{1}{c}{0.47} \\
\cline{2-9}          & \multicolumn{1}{c||}{GC-NET\,\cite{kendall17}} & \multicolumn{1}{c|}{\textbf{2.21}} & \multicolumn{1}{c|}{6.16} & \multicolumn{1}{c||}{2.87} & \multicolumn{1}{c|}{\textbf{2.02}} & \multicolumn{1}{c|}{5.58} & \multicolumn{1}{c||}{2.61} & \multicolumn{1}{c}{0.9} \\
\cline{2-9}          & \multicolumn{1}{c||}{DRR\,\cite{gidaris17}} & \multicolumn{1}{c|}{2.58} & \multicolumn{1}{c|}{6.04} & \multicolumn{1}{c||}{3.16} & \multicolumn{1}{c|}{2.34} & \multicolumn{1}{c|}{4.87} & \multicolumn{1}{c||}{2.76} & \multicolumn{1}{c}{0.4} \\
\cline{2-9}          & \multicolumn{1}{c||}{L-ResMatch\,\cite{shaked17}} & \multicolumn{1}{c|}{2.72} & \multicolumn{1}{c|}{6.95} & \multicolumn{1}{c||}{3.42} & \multicolumn{1}{c|}{2.35} & \multicolumn{1}{c|}{5.74} & \multicolumn{1}{c||}{2.91} & \multicolumn{1}{c}{48*} \\
\cline{2-9}          & \multicolumn{1}{c||}{Displets\,v2\,\cite{guney15}} & \multicolumn{1}{c|}{3.00} & \multicolumn{1}{c|}{5.56} & \multicolumn{1}{c||}{3.43} & \multicolumn{1}{c|}{2.73} & \multicolumn{1}{c|}{4.95} & \multicolumn{1}{c||}{3.09} & \multicolumn{1}{c}{265*} \\
\cline{2-9}          & \multicolumn{1}{c||}{D3DNet} & \multicolumn{1}{c|}{2.88} & \multicolumn{1}{c|}{6.60} & \multicolumn{1}{c||}{3.50} & \multicolumn{1}{c|}{2.71} & \multicolumn{1}{c|}{6.08} & \multicolumn{1}{c||}{3.26} & \multicolumn{1}{c}{\textbf{0.35}} \\
\cline{2-9}          & \multicolumn{1}{c||}{SsSMNet} & \multicolumn{1}{c|}{2.86} & \multicolumn{1}{c|}{7.12} & \multicolumn{1}{c||}{3.57} & \multicolumn{1}{c|}{2.63} & \multicolumn{1}{c|}{6.26} & \multicolumn{1}{c||}{3.23} & \multicolumn{1}{c}{0.8} \\
\cline{2-9}          &       &       &       &       &       &       &       &  \\
    \end{tabular}%
  \caption{Leading submissions of the KITTI 2015 stereo online leaderboard (as of August 2017). Three-pixel-error of our approach and the other state-of-the-art methods are tabulated, where our approach ranks first. The symbol ``*'' denotes runtime on CPU.}
  \label{tab:res_kit}%
\end{table*}%

\subsection{Discussions}
Existing end-to-end CNNs for stereo matching, {\it e.g.}, \cite{kendall17,mayer16} and this work, all relies on a vast amount of training data with ground-truth. 
However, it is costly to collect depth data in the real physical world; while synthetic data, {\it e.g.}, the FlyingThings3D dataset, cannot fully reflects the properties of the real environment.

A potential solution to the above dilemma is to borrow the wisdom from traditional approaches and embed the left-right consistency check module into the CNNs.
As mentioned in Section\,\ref{sec:related}, it is explored by  \cite{godard17,kuznietsov17} for monocular depth estimation, leading to unsupervised (or semi-supervised) method requiring (very) little amount of data with ground-truth. 
However, recent end-to-end CNN-based approaches already produces very accurate disparity estimates, in contrast to the case of monocular depth estimation.
As a result, any new mechanisms ({\it e.g.}, left-right consistency check in this case) introduced to the networks need to be very reliable/robust, otherwise further improvements cannot be achieved. 
We leave this problem of designing robust left-right consistency check module for future investigation.

\section{Conclusions}
\label{sec:conclude}
Recent works employing CNNs for stereo matching have achieved prominent performance. 
Nevertheless, estimating high-quality disparity for inherently ill-posed regions remains intractable. 
In this work, we propose a cascade CNN architecture with two stages: the first stage manages to produce an initial disparity image with fine details, while the second stage explicitly refines/rectifies the initial disparity with residual signals across multiple scales. 
We call our approach cascade residual learning. 
Our experiments show that, residual learning not only provides effective refinement but also benefits the optimization of the whole two-stage network. 
Our approach achieves state-of-the-art stereo matching performance, it ranks first in the KITTI 2015 stereo benchmark, exceeding the prior works by a noteworthy margin.

{\small
\bibliographystyle{ieee}
\bibliography{ref}
}

\end{document}